\pdfoutput=1

\documentclass[11pt]{article}

\usepackage{acl}
\usepackage{listings}
\usepackage{amsmath}
\usepackage{algorithm}
\usepackage{algpseudocode}
\makeatletter
\renewcommand{\ALG@name}{Prompt} 
\makeatother
\usepackage{times}
\usepackage{latexsym}
\usepackage[utf8]{inputenc}

\usepackage[T1]{fontenc}

\usepackage[utf8]{inputenc}

\usepackage{microtype}

\usepackage{inconsolata}

\usepackage{graphicx}

\usepackage{booktabs}
\usepackage{xcolor}
\usepackage{colortbl}
\usepackage{adjustbox}
\usepackage{stfloats}

\title{To Bias or Not to Bias: Detecting bias in News with bias-detector}

\author{
  \textbf{Himel Ghosh\textsuperscript{1,2}},
  \textbf{Ahmed Mosharafa\textsuperscript{1}},
  \textbf{Georg Groh\textsuperscript{1}}
\\
  \textsuperscript{1}Technical University of Munich (TUM), Germany,
\\
  \textsuperscript{2}Sapienza University of Rome, Italy,
\\
  \small{
    \textbf{Email:} \href{himel.ghosh@tum.de}{himel.ghosh@tum.de}
  }
}

\begin{document}
\pagestyle{plain}
\maketitle
\begin{abstract}
Media bias detection is a critical task in ensuring fair and balanced information dissemination, yet it remains challenging due to the subjectivity of bias and the scarcity of high-quality annotated data. In this work, we perform sentence-level bias classification by fine-tuning a RoBERTa-based model on the expert-annotated BABE dataset. Using McNemar's test and the 5×2 cross-validation paired t-test, we show statistically significant improvements in performance when comparing our model to a domain-adaptively pre-trained DA-Roberta-FT baseline. Furthermore, attention-based analysis shows that our model avoids common pitfalls like oversensitivity to politically charged terms and instead attends more meaningfully to contextually relevant tokens. For a comprehensive examination of media bias, we present a pipeline that combines our model with an already-existing bias-type classifier.  Our method exhibits good generalization and interpretability, despite being constrained by sentence-level analysis and dataset size because of a lack of larger and more advanced bias corpora. We talk about context-aware modeling, bias neutralization, and advanced bias type classification as potential future directions. Our findings contribute to building more robust, explainable, and socially responsible NLP systems for media bias detection.

\end{abstract}


\section{Introduction}
Journalism is supposed to be neutral and free from biases, but that is often not followed. Cambridge English dictionary defines bias as "the action of supporting or opposing a particular person or thing in an unfair way, because of allowing personal opinions to influence your judgment". Intentional or unconscious, media bias is the unfair treatment of particular individuals and ideas as well as the reporting of certain biased viewpoints \cite{mediabias} which can significantly influence public opinion, skew political and democratic discourses.

For a free and fair journalism and news reporting, it is imperative to identify the bias and eventually mitigate them. This paper addresses the ever-increasing need for bias identification models and demonstrates their statistical significance compared to the other available models . Here we introduce a neural transformer-based fine-tuned bias detection model trained on Bias Annotations by Experts (BABE) dataset by \cite{Spinde_2021} for robust bias detection available on \href{https://huggingface.co/himel7/bias-detector}{Hugging Face}.

\begin{figure}
    \centering
    \includegraphics[width=1\linewidth]{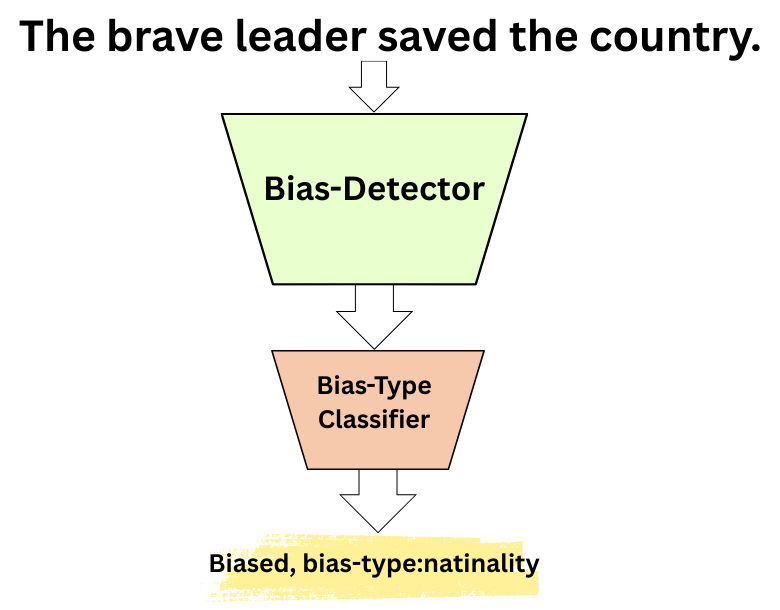}
    \caption{Bias Detection and Classification Scheme which shows how the pipeline detects sentence-level biases.}
    \label{fig:scheme}
\end{figure}

Furthermore, we present a pipeline (See Fig. \ref{fig:scheme}) with both bias detection and bias classification for bias analysis on sentences. These findings will offer a framework for future research in the media bias detection.

\section{Related Work}
At the nexus of political science, psychology, and natural language processing, media bias detection has emerged as a key field of study. Historically, research in this field has used datasets like AllSides, MediaBias/FactCheck, and Media Frames Corpus to identify ideological bias at the source or article level. Lexicon-based heuristics or shallow machine learning models that used manually constructed features like sentiment polarity, assertive verbs, or hedge words were frequently used in early computational approaches \cite{recasens-etal-2013-linguistic,hube}.

However, due to their low inter-annotator agreement and dependence on small datasets, these approaches showed limitations in terms of scalability and generalization. For instance,\cite{lim-etal-2020-annotating,farber} used crowdsourcing to create datasets but found low interrater reliability (e.g., Krippendorff's $\alpha$ = 0.0), indicating challenges in capturing the intrinsically subjective nature of bias.

To address these challenges, Spinde et al. introduced the MBIC and BABE datasets— two of the most influential resources in recent media bias research. MBIC included sentence- and word-level annotations with detailed annotator background metadata \cite{spinde2021mbicmediabias}. BABE (Bias Annotations By Experts), built by trained annotators, significantly improved annotation quality, offering around 4000 high-quality labeled sentences across topics and biases \cite{Spinde_2021}.

Regarding modeling, previous approaches used logistic regression or random forests based on manually created linguistic features \cite{spinde2020}. With the introduction of attention in transformers by \cite{NIPS2017}, later work adopted transformer-based models, especially BERT, RoBERTa, and XLNet, for sentence-level classification \cite{devlin-etal-2019-bert,liu2019robertarobustlyoptimizedbert}. These models showed improved contextual sensitivity and generalization. With noisy but scalable training data, distant supervision has become a promising method for producing weak labels from partisan sources (like AllSides) \cite{tang-etal-2014-learning,Spinde_2021}.

A significant recent contribution to bias detection is made by \cite{Krieger_2022}, who introduced DA-Roberta-FT, a domain-adapted RoBERTa model pre-trained on the Wiki Neutrality Corpus (WNC) \cite{pryzant2019automaticallyneutralizingsubjectivebias} and fine-tuned on the BABE dataset. Although their method achieves a new state-of-the-art F1 score, our approach which is developed based on their work, achieves a higher score and subsequently proved consistent in statistical signifance testing.

\cite{powers2025} recently contributed to the classification of several types of biases such as religious bias, racial bias, political bias, and others through their work on GUS framework and published a \href{https://huggingface.co/maximuspowers/bias-type-classifier}{bias-type-classifier} model which forms an integral part of the bias-analysis pipeline of our work.


Our work contributes to this gap by:

\begin{enumerate}
    \item \textbf{Adapting and fine-tuning} a RoBERTa-base model on BABE dataset thereby providing a robust bias detector.
    \item \textbf{Statistical Significance Testing} of our proposed model to establish its ground as a significant enhancement to the previous models.
    \item Integrating with a \textbf{bias-type classifier} to reveal granular categories of the biases.
\end{enumerate}
Our contribution provides a robust bias detector and classifier by fusing neural modeling, domain-specific fine-tuning, and adapting to existing methodology.

\section{Methodology}
\subsection{Model Selection and Setup}
Our goal is to develop a robust model for sentence-level bias classification in news articles. We build upon prior work by \cite{Krieger_2022}, who introduced DA-Roberta-FT, a RoBERTa-base model \cite{xlmroberta} pre-trained on the Wiki Neutrality Corpus (WNC) introduced by \cite{pryzant2019automaticallyneutralizingsubjectivebias} for domain adaptation and fine-tuned on the BABE dataset introduced by \cite{Spinde_2021}. The dataset has a split of 3.12k training instances and 1k test instances. Their results demonstrated the effectiveness of domain-specific pre-training over general-purpose models for bias detection.

Inspired by their setup, we evaluate DA-RoBERTa-FT as a baseline and use it as a reference point for our model selection and evaluation. We tested their \href{https://huggingface.co/mediabiasgroup/da-roberta-babe-ft}{DA-Roberta-FT model} with the test split of the BABE Dataset yielding the best macro F1 score of 0.823.
To enhance this, we create another model by taking the transformer \cite{NIPS2017} model: \href{https://huggingface.co/FacebookAI/roberta-base}{Roberta base} \cite{roberta} and fine-tuned it directly on the train split of the BABE dataset. From this train split, we reserve 10\% of the instances as a validation set, stratified by class. Early stopping is applied based on macro-F1 on the validation set. Once the best checkpoint is selected, the model is frozen and evaluated exactly once on the official BABE test split (1k sentences) and achieved the best macro F1 score of 0.8517.

\begin{figure}
    \centering
    \includegraphics[width=1\linewidth]{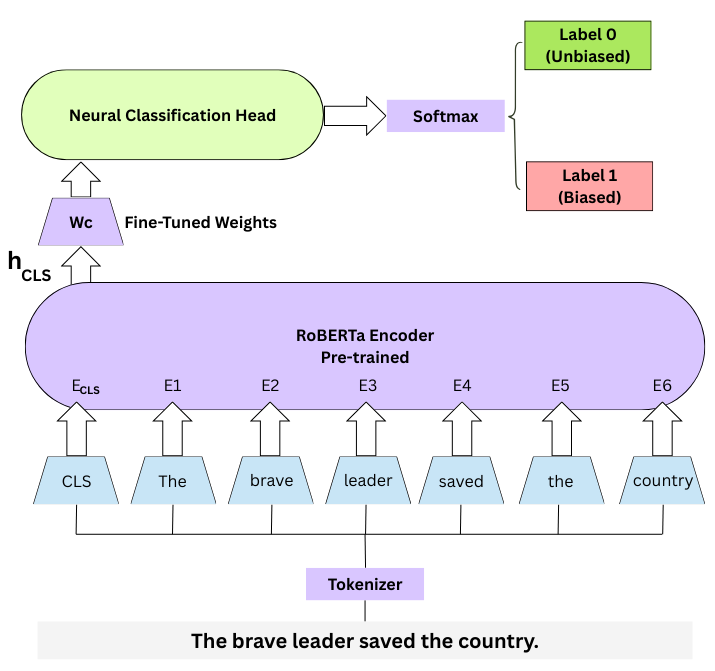}
    \caption{Architecture of the Bias Detector Model. The output vector of the CLS token serves as input to the classifier head}
    \label{fig:architecture}
\end{figure}





\subsection{Learning Task and Objective}

We frame sentence-level bias detection as a binary supervised classification problem.
Given a sentence $x_{i} \in X$, the objective is to assign a label $y_{i} \in {0,1}$, where:

\begin{itemize}
\item $y_i = 1$ denotes a biased sentence
\item $y_i = 0$ denotes an unbiased sentence
\end{itemize}

Let $D = {(x_i, y_i)}_{i=1}^{N}$ denote the training set. We fine-tune a RoBERTa-base encoder with a two-class softmax classification head.
For each input $x_i$, RoBERTa produces a contextual representation of the initial sentence boundary token (conventionally referred to as “CLS”, corresponding to RoBERTa’s $<s>$ token).
This vector is passed through a linear layer $W_c$ to produce a pair of logits:
\begin{equation}
z_i = W_ch_{CLS,i}
\end{equation}
The model predicts a probability distribution over the two labels via a softmax:
\begin{equation}
\hat{y_i} = softmax(z_i)
\end{equation}

where $\hat{y}_{i,1}$ is the predicted probability that $x_i$ is biased.

Since the classifier outputs two logits, we optimise the standard multi-class cross-entropy loss:

\begin{equation}
\mathcal{L} = -\frac{1}{N}\sum_{i=1}^{N} \log \hat{y}_{i, y_i}
\end{equation}

This objective jointly trains the encoder and classification head end-to-end during fine-tuning.

To complement binary bias detection with bias-type attribution, we additionally incorporate the model by \cite{powers2025}, released as the \texttt{bias-type-classifier} on the Hugging Face Hub.\footnote{\url{https://huggingface.co/maximuspowers/bias-type-classifier}}
Together, these components form a complete pipeline: given an input sentence, the system determines whether it is biased and, if so, identifies the specific category of bias expressed.

\subsection{Evaluation}

To ensure fair and replicable comparison with DA-Roberta-FT, we adopted the same evaluation protocol as \cite{Krieger_2022}, namely:
To ensure a fair and reproducible comparison with DA-RoBERTa-FT \cite{Krieger_2022}, we re-evaluated both models using the official BABE test split (3.12k/1k). Hence we used the following:
\begin{itemize}
    \item \textbf{Dataset:} BABE (Bias Annotations By Experts), consisting of 4120 sentences annotated by expert annotators with 3120 training and 1000 test cases.
    \item \textbf{Evaluation Metric:} Macro F1 score, reported with standard error over splits.
\end{itemize}

This ensures that both models are evaluated under exactly the same conditions, using identical data, preprocessing, and metric computation, and avoids discrepancies arising from different fold construction, random seeds, or training procedures.
The observations are further established by the following:

\textbf{Statistical Significance Testing}:
\begin{itemize}
    \item \textbf{McNemar's Test:} To measure the statistical significance of the differences, following \cite{hitchhikers}, we conducted McNemar's Test. According to the test's null hypothesis, both algorithms have the same marginal probability for each outcome (label zero or label one).
    \item \textbf{5x2 CV Test:}To further establish the evidence, we performed the more robust 5x2 CV paired t-test according to \cite{thomas}.
\end{itemize}

Our code is available on this \href{https://github.com/Himel1996/NewsBiasDetector/}{Github Repository} and the model is available on this \href{https://huggingface.co/himel7/bias-detector}{link}.

\section{Experiments}

We evaluate the DA-Roberta-FT model by \cite{Krieger_2022} using Hugging Face’s implementation and load their model from the official release. Our model: \href{https://huggingface.co/himel7/bias-detector}{bias-detector} is fine-tuned using the Hugging Face Transformers library on the full BABE dataset. We used the AdamW optimizer with a learning rate of 2 x $10^{-5}$, batch size 32, and early stopping based on validation F1. All experiments were run on RTX A6000 GPU. Fine tuning took around ~30 minutes. The best model checkpoint was then published on Hugging face library. Both models were put through the above-mentioned evaluation strategy to achieve the macro F1 scores and statistical scores.

Similar to \cite{Krieger_2022}, we do not know the underlying distribution of our target metric and \cite{hitchhikers} suggests us to choose McNemar's test. It is a non-parametric statistical test used to compare the performance of two classifiers on the same data instances, focusing specifically on paired nominal data — i.e., whether each model was correct or incorrect on each example. It is based on a 2x2 contingency table highlighting the model prediction on n- data points of the test set. The following are the hypotheses, respectively:

\noindent\textbf{McNemar's Test Null Hypothesis ($H_0$):}

$P(\text{Model A correct, Model B incorrect})$ = $P(\text{Model B correct, Model A incorrect})$

\noindent\textbf{Alternative Hypothesis ($H_1$):}

$P(\text{Model A correct, Model B incorrect})$$ \neq P(\text{Model B correct, Model A incorrect})$

The test statistic follows a $\chi^2$ distribution with 1 degree of freedom. If $\chi^2$ is large and p-value<0.05, we reject $H_0$ and conclude that the models' performances are significantly different.

To assess whether the difference in performance between our model and the baseline is statistically significant, we employed the 5x2 cross-validation paired t-test \cite{thomas}. This test runs two models over five independent 2-fold cross-validation splits and computes a t-statistic over the difference in performance (e.g., F1-score) across these trials. The test statistic is based on 10 independent estimates of the performance difference.
The null hypothesis assumes that both models have equal expected performance.

\noindent\textbf{5x2 cv test Null Hypothesis ($H_0$):}
\begin{equation}
E[\theta_i] = 0
\end{equation}

\noindent\textbf{Alternative Hypothesis ($H_1$):}
\begin{equation}
E[\theta_i] \neq 0
\end{equation}
Where $\theta_i$ is the difference in performance (e.g., accuracy or F1-score) between the two models on the ith fold. A statistically significant result (p < 0.05) indicates that the observed difference is unlikely to be due to random chance.

\section{Results}
Our model outperforms the DA-Roberta-FT baseline in terms of macro F1 score.

Adopting their research and adding to the Table 1 of \cite{Krieger_2022}, we present the updated Table \ref{tab:res} here with our model's performance on macro F1 and standard error in parentheses.

\begin{table}
    \centering
    \begin{tabular}{cc}
        \textbf{Model} & \textbf{Macro F1 (error)}\\
        \hline
         BERT& 0.789 (0.011) \\
         RoBERTa& 0.799 (0.011)\\
         DA-Roberta-FT& 0.823 (0.004)\\
         Bias-Detector& \textbf{0.852} (0.003)\\
         \hline
    \end{tabular}
    \caption{Evaluation Results: Bias Detector outperforms the state of the art baseline with higher macro F1 score on the same evaluation setup.}
    \label{tab:res}
\end{table}

Furthermore, we conducted McNemar’s test to assess the significance of the performance difference between our model (bias-detector) and the DA-Roberta-FT baseline. The test yielded a p-value of 0.0167, indicating a statistically significant improvement in classification performance by our model which supports the observed difference in the macro-F1 scores. Table \ref{tab:mcnemar_results} shows the McNemar Test result comparing our model with the baseline.

\begin{table}[h]
\centering
\begin{tabular}{|c|c|}
\hline
\textbf{Chi-squared ($\chi^2$)} & \textbf{p-value}\\
\hline
5.72 & 0.0167 \\
\hline
\end{tabular}
\caption{McNemar’s Test Results comparing bias-detector and DA-RoBERTa-FT on the BABE dataset test split. p-value indicates statistically significant difference at $\alpha = 0.05$.}
\label{tab:mcnemar_results}
\end{table}

\begin{figure}
    \centering
    \includegraphics[width=1\linewidth]{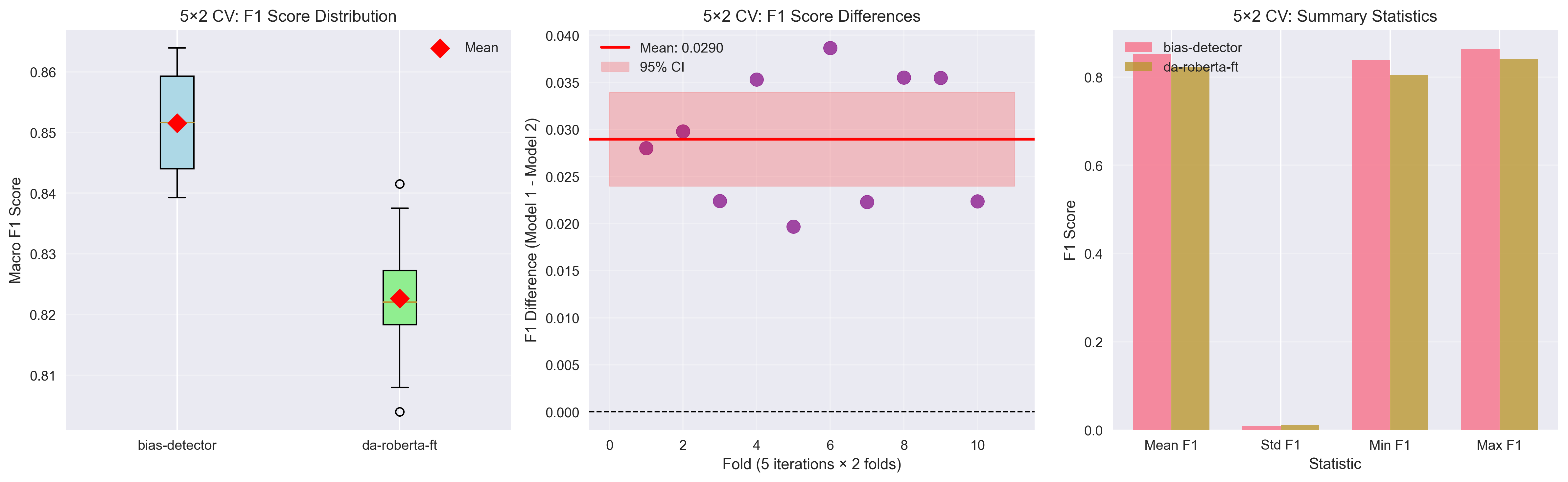}
    \caption{5x2 CV t-test Results: Statistically significant difference on Macro-F1 score across 5 folds of the CV t-test between bias-detector and da-roberta-ft baseline}
    \label{fig:cvt}
\end{figure}

Based on \cite{thomas}, 5x2 CV t-test is slightly more powerful than McNemar's test, we conducted the 5×2 cross-validation paired t-test to assess whether the observed improvement in our model over the DA-Roberta-FT baseline is statistically significant. The test yielded a t-statistic of 13.12 with a p-value of 3.59 x $10^{-7}$ confirming that the performance difference is statistically significant. See Fig. \ref{fig:cvt} for reference.

These results confirm that the observed gains are not due to random variation but reflect a consistent enhancement in bias detection accuracy with high statistical significance.

\subsection{Interpretation}
\begin{figure}
    \centering
    \includegraphics[width=1\linewidth]{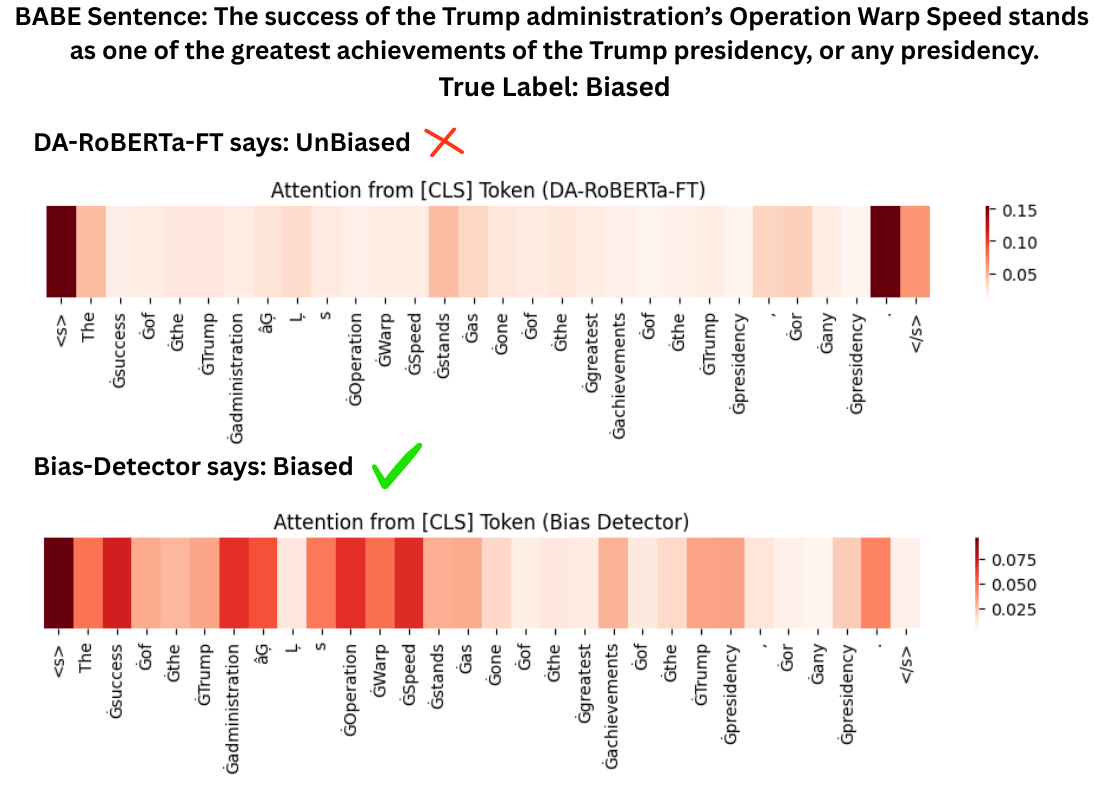}
    \caption{Attention weights heatmap for False Negatives}
    \label{fig:attnFN}
\end{figure}

\begin{figure}
    \centering
    \includegraphics[width=1\linewidth]{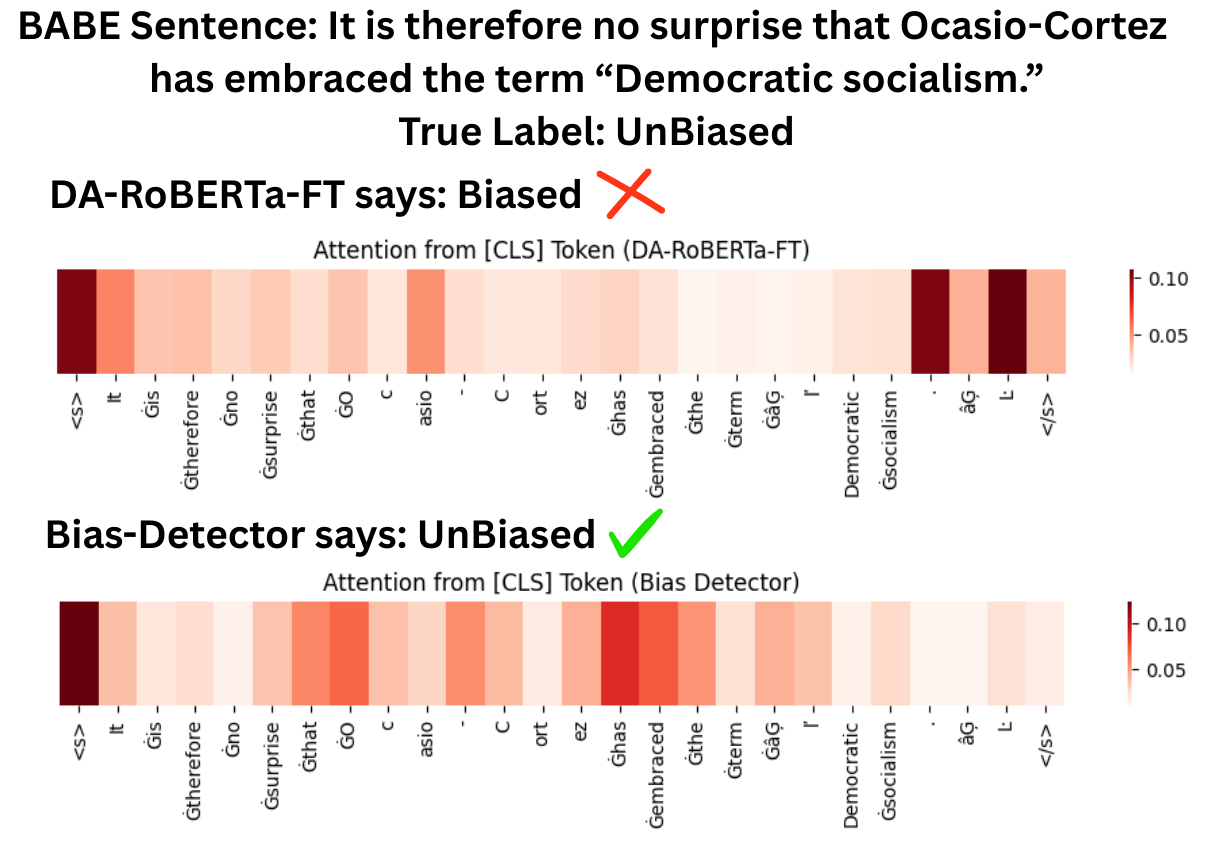}
    \caption{Attention weights heatmap for False Positives}
    \label{fig:attnFP}
\end{figure}

While DA-Roberta-FT provides many false positives (unbiased sentence flagged as biased) and false negatives (biased sentence flagged as unbiased), our bias-detector proved to be more robust. Check Fig. \ref{fig:attnFN} and \ref{fig:attnFP} to see how Bias-Detector attends to the tokens compared to the DA-Roberta-FT model.

As you see, in Fig. \ref{fig:attnFN}, the baseline DA-Roberta-FT does not attend to the tokens contributing to bias. It is an example of framing bias missed by DA-Roberta-FT. The sentence uses strong evaluative language (e.g., “success”, “greatest achievements”) to frame a political action positively. DA-Roberta-FT fails to detect this bias possibly because this domain-adapted model may be biased toward formal neutrality, especially if WNC was too Wikipedia-like and underexposed to superlative media framing; while our model correctly predicts "biased" and focuses attention on key framing tokens because of better linguistic generalization.

In other cases, like Fig. \ref{fig:attnFP}, illustrates DA-Roberta-FT incorrectly classifies a neutral sentence as biased due to over-attention to politically charged tokens like “Democratic socialism.” This reflects domain-adaptive overfitting — the model may have learned to associate certain political keywords with bias regardless of context. In contrast, our model correctly classifies the sentence as unbiased and attends more meaningfully to contextual phrases like “has embraced.” This suggests that our bias-detector generalizes better by avoiding lexical bias triggers and relying on contextual framing. 

\subsection{Discussion}

The DA-Roberta-FT model took a two-step approach: first training on  Wiki-Neutrality Corpus (WNC, Wikipedia-style content) using a multilingual model base (xlm-roberta-base) for domain-adaptation, then fine-tuning for the task. In contrast, our model skipped that first step and went straight to fine-tuning a standard English RoBERTa model (roberta-base) on the BABE dataset which is focused on the news-domain.
Our approach worked better for a few key reasons. First, our model used a smaller, English-focused foundation rather than a larger multilingual one, which meant it was already well-suited for English text processing. Hence it generalized well to the linguistic subtleties of English language. Second, the vocabulary and how words are broken down into tokens were better matched to the task at hand. Finally, by training directly on news articles (the BABE dataset) rather than first adapting to Wikipedia-style content, our model avoided the potential confusion that comes from the domain-shift, that is, switching between different types of text.
All of these factors likely helped our model become better at spotting the subtle language patterns that indicate bias at the sentence level - exactly what we needed it to do.
Hence, our work provides a more robust state-of-the-art bias-detection model available on Hugging Face library which can be widely used for bias identification and news bias analysis by media bias researchers, NLP scientists and the likes. We also laid down the framework for further bias analysis by integrating an existing bias type classifier model in the same pipeline of our bias detection, thereby opening up new avenues of research work in this zone.

\section{Limitations and Future Works}
Although our bias detection model demonstrates better performance over existing baselines, it is inherently constrained by the limitations of the dataset it was trained on. Our work is restricted by the limitations of The BABE dataset which comprises approximately of 4,000 English-language sentences sourced from several media outlets, which limits both the linguistic and topical diversity of training examples. Language is ever growing and definition of bias is also subjective. The question of who defines what as biased and whether something would be considered as a bias in a specific socio-cultural context or not, is still somewhat perspective dependent. Hence our model is also bounded by the language and data limitations besides being a sentence level model which only classifies within 2 labels with limited interpretability. This limits the model's ability to generalize to paragraph-level or document-level bias and may restrict applicability to more complex real-world media scenarios.

As part of the future scope, addressing these limitations shall be the core focus. Trying to increase the interpretability by extending the action on each of  token paragraph or document level can be a way forward. Also the need for a larger, more diverse corpus beyond BABE will spearhead the development. Evaluating and adapting the model to other media domains (e.g., blogs, social media) and non-English languages could provide insights into how bias manifests across cultures and platforms. Future work could explore richer interpretability techniques such as SHAP, integrated gradients, or counterfactual examples. These can be integrated into interactive tools for journalists, fact-checkers, or researchers to manually inspect or override model predictions.
Beyond identification and classification, generative approaches could be used to neutralize biased content while preserving factual meaning. Combining classification with generation (e.g., using T5 or LLMs) could enable controllable text rewriting for media editing or curation purposes.

We believe these directions will advance the development of more robust, explainable, and socially responsible NLP systems for bias detection in media.

\section{Conclusion}

In this work, we presented a fine-tuned RoBERTa-based model called \textbf{bias-detector} for sentence-level media bias detection, trained on the expert-annotated BABE dataset. Through comprehensive evaluation and statistical testing, our model outperformed the domain-adapted DA-RoBERTa-FT baseline across multiple folds, achieving higher macro F1-score and demonstrating statistically significant enhancements. Attention-based analyses further revealed that our model attends more meaningfully to linguistic cues associated with bias.

Our findings suggest that careful fine-tuning of the rightly chosen base model on task-specific data, even without domain-adaptive pretraining, can yield robust and interpretable models for media bias detection. We incorporated an already available bias-type-classifier in our pipeline to show a framework for complete bias analysis. We also highlight the importance of aligning model design with linguistic and contextual nuances inherent in bias perception.

While our approach is constrained by dataset size, language scope, and only sentence level analysis, it opens the door to several new vistas. These include fine-grained bias type classification, discourse-level modeling, multilingual adaptation, and bias-neutralization through natural language generation. We hope our work contributes to the development of more transparent, context-aware, and socially responsible NLP systems for media analysis.

\section{Acknowledgements}
I acknowledge the support and guidance of my Master-degree thesis supervisor Ahmed Mosharafa and Professor Georg Groh.

\bibliography{custom}


\end{document}